\documentclass[10pt,twocolumn,letterpaper]{article}

\usepackage[pagenumbers]{cvpr} % To force page numbers, e.g. for an arXiv version
\usepackage{multirow}

\definecolor{cvprblue}{rgb}{0.21,0.49,0.74}
\usepackage[pagebackref,breaklinks,colorlinks,allcolors=cvprblue]{hyperref}

\def\paperID{1233} % *** Enter the Paper ID here
\def\confName{CVPR}
\def\confYear{2025}

\title{AlignMamba: Enhancing Multimodal Mamba with Local and Global Cross-modal Alignment}

\author{
Yan Li$^{1}$, Yifei Xing$^{1}$, Xiangyuan Lan$^{1}$, Xin Li$^{1}$, Haifeng Chen$^{2}$, Dongmei Jiang$^{1*}$\\
$^{1}$Pengcheng Laboratory, Shenzhen, China\\
$^{2}$Shaanxi University of Science and Technology, Xi'an, China\\
{
\tt\small 
liyan4ai@gmail.com,
\{xingyf, lanxy, lix07, jiangdm\}@pcl.ac.cn,
chenhaifeng@sust.edu.cn}
}

\begin{document}
\maketitle
\begin{abstract}
Cross-modal alignment is crucial for multimodal representation fusion due to the inherent heterogeneity between modalities. While Transformer-based methods have shown promising results in modeling inter-modal relationships, their quadratic computational complexity limits their applicability to long-sequence or large-scale data. Although recent Mamba-based approaches achieve linear complexity, their sequential scanning mechanism poses fundamental challenges in comprehensively modeling cross-modal relationships. To address this limitation, we propose AlignMamba, an efficient and effective method for multimodal fusion. Specifically, grounded in Optimal Transport, we introduce a local cross-modal alignment module that explicitly learns token-level correspondences between different modalities. Moreover, we propose a global cross-modal alignment loss based on Maximum Mean Discrepancy to implicitly enforce the consistency between different modal distributions. Finally, the unimodal representations after local and global alignment are passed to the Mamba backbone for further cross-modal interaction and multimodal fusion. Extensive experiments on complete and incomplete multimodal fusion tasks demonstrate the effectiveness and efficiency of the proposed method. For instance, on the CMU-MOSI dataset, AlignMamba improves classification accuracy by 0.9\%, reduces GPU memory usage by 20.3\%, and decreases inference time by 83.3\%.
\end{abstract}
\section{Introduction}
In recent years, multimodal representation fusion has emerged as a critical technology for integrating and understanding information across different modalities (e.g., audio, video, language). This capability is fundamental to a wide range of applications such as visual-language understanding~\cite{zhang2023multimodal} and audio-visual analysis~\cite{zadeh2018multimodal,lee2020cross}. However, due to the inherent heterogeneity between modalities - each with its distinct statistical properties and feature distributions - achieving effective cross-modal alignment and fusion remains a significant challenge.

Traditional approaches to this challenge have primarily relied on Transformer-based~\cite{vaswani2017attention} architectures, which can be broadly categorized into two paradigms. Single-stream methods (e.g., VisualBERT~\cite{li2019visualbert}, ViLT~\cite{kim2021vilt}, LLaVA~\cite{liu2023visual}) concatenate features from different modalities into a unified sequence and process them through a shared Transformer layer. In contrast, multi-stream approaches (e.g., LXMERT~\cite{tan2019lxmert}, ViLBERT~\cite{lu2019vilbert}, MulT~\cite{tsai2019multimodal}, CMA~\cite{zheng2022multi}) employ separate encoders for each modality with cross-modal Transformers to facilitate information exchange. While these methods have demonstrated promising results in capturing dynamic cross-modal interactions, they suffer from a fundamental limitation: the quadratic computational complexity of attention mechanisms makes them inefficient for processing long-sequence or large-scale data common in real-world multimodal applications.

\begin{figure}[t]
\centering
\includegraphics[width=\linewidth]{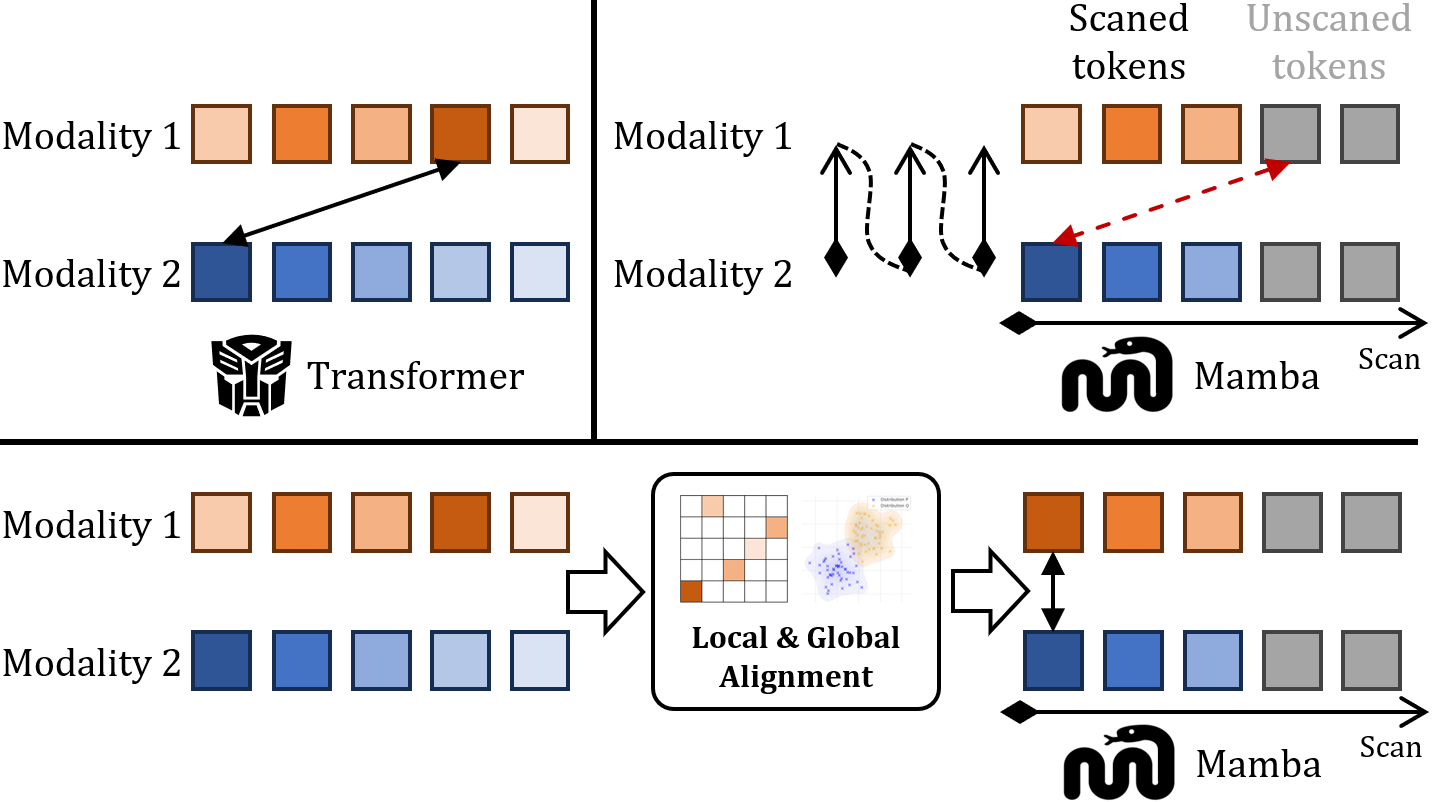}
\caption{Transformer leverages attention mechanisms to model relationships across different modalities (top left), whereas Mamba struggles to achieve this due to its sequential scanning mechanism (top right). In contrast, the proposed AlignMamba utilizes both local (OT-based) and global (MMD-based) cross-modal alignment information to achieve efficient and effective multimodal fusion (bottom).}
\label{fig:motivation}
\end{figure}

Recent advances in sequence modeling have introduced the Mamba~\cite{gu2023mamba} architecture, based on State Space Models (SSMs)~\cite{gu2021combining,gu2022parameterization}, which achieves linear computational complexity while maintaining strong modeling capabilities. By incorporating selection mechanisms and hardware-aware parallel algorithms into SSMs, Mamba effectively captures long-range dependencies without the computational burden of attention mechanisms. This breakthrough has sparked considerable interest in adapting Mamba for multimodal fusion tasks, with approaches ranging from direct feature concatenation (e.g., VL-Mamba~\cite{qiao2024vl}, Cobra~\cite{zhao2024cobra}, RoboMamba~\cite{liu2024robomamba}) to multi-stream architectures (e.g., Pan-Mamba~\cite{he2024pan}, Fusion-Mamba~\cite{dong2024fusion}, MambaDFuse~\cite{li2024mambadfuse}). However, our analysis reveals a critical limitation. As shown in Fig.~\ref{fig:motivation}, Mamba's sequential scanning mechanism, while computationally efficient, struggles to capture comprehensive cross-modal relationships, particularly with unscanned tokens. This inherent limitation leads to suboptimal alignment between modalities and consequently affects the quality of learned multimodal fusion representations.

To address these issues, we propose AlignMamba, which integrates local and global cross-modal alignment information into Mamba for efficient and effective multimodal fusion. Specifically, we introduce a local alignment module based on Optimal Transport (OT), which learns a transport plan to align features across different modalities by minimizing the cost of feature transportation. While local alignment captures token-level cross-modal relationships, it does not account for distributional differences between modalities. Therefore, we also propose a global alignment loss based on Maximum Mean Discrepancy (MMD). Leveraging the theoretical advantages of reproducing kernel Hilbert space, MMD maps the feature distributions of different modalities into a high-dimensional space and achieves implicit alignment by minimizing their distributional differences. After local and global cross-modal alignment, all unimodal features are combined and fed into the Mamba backbone for further multimodal fusion. This dual alignment strategy ensures that Mamba can exploit both local and global relationships between modalities, thereby learning more comprehensive multimodal representations.

In summary, the contributions of this paper are threefold:
\begin{itemize}
\item We observe the limitation of directly applying Mamba to multimodal fusion tasks, which ignores more comprehensive cross-modal alignment information, and propose the AlignMamba framework to achieve efficient and effective multimodal fusion.
\item We introduce an OT-based local alignment module for explicit learning of token-level correspondences, complemented by an MMD-based global alignment loss for implicit distribution alignment. These two types of alignment information complement each other, achieving comprehensive cross-modal alignment.
\item Extensive experiments on both complete and incomplete multimodal fusion tasks demonstrate that AlignMamba achieves state-of-the-art results in terms of both effectiveness and efficiency.
\end{itemize}
\section{Related work}
\subsection{Transformer-based Multimodal Fusion}
Transformer~\cite{vaswani2017attention}, with its powerful modeling capabilities, has become the cornerstone architecture in modern neural networks. Existing multimodal fusion methods mainly rely on Transformers to model relationships between different modalities and learn multimodal fusion representations. These approaches can be categorized into two main types: multi-stream and single-stream methods.

Multi-stream methods employ cross-modal Transformers to model interactions between any two modalities. For vision-language pre-training tasks, models like ViLBERT~\cite{lu2019vilbert} and LXMERT~\cite{tan2019lxmert} utilize two co-attention Transformer layers to model bidirectional relationships between visual and textual modalities. For audio-visual-textual trimodal fusion tasks, MulT~\cite{tsai2019multimodal} leverages cross-modal Transformers to model pairwise modal interactions, and then concatenate all bimodal fusion representations to obtain trimodal fusion representations. Similarly, CMA~\cite{tsai2019multimodal}, based on cross-modal attention mechanisms, was proposed to fuse features from three modalities. More recently, BLIP-2~\cite{li2023blip} introduced Q-Former, a lightweight querying Transformer architecture, to align vision-language modalities and learn multimodal fusion representations.

Single-stream methods adopt a more straightforward strategy by concatenating features from different modalities and feeding them into a Transformer encoder for cross-modal interaction and multimodal fusion. For instance, in vision-language pre-training tasks, VisualBERT~\cite{li2019visualbert} extracts features from key regions using object detectors and concatenates these region feature sequences with text token embeddings before feeding them into a Transformer. In contrast, ViLT~\cite{kim2021vilt} replaces region feature sequences with image patch embedding sequences, discarding the object detection backbone and improving efficiency. Recent multimodal pre-training models, such as LLaVA~\cite{liu2023visual}, have adopted similar approaches to model cross-modal correspondences and learn multimodal fusion representations for downstream tasks.

Existing methods achieve cross-modal interaction and fusion through cross-attention or self-attention mechanisms, learning comprehensive and effective multimodal fusion representations. However, the quadratic time complexity of Transformers limits their efficiency when processing large-scale or long-sequence data. This limitation necessitates the development of novel multimodal fusion methods that balance effectiveness and efficiency.

\subsection{Mamba-based Multimodal Fusion}
As a novel architectural paradigm, Mamba~\cite{gu2023mamba} incorporates selection mechanisms and hardware-aware parallel algorithms into SSMs~\cite{gu2021combining,gu2022parameterization}, achieving efficient and effective sequence modeling in the language domain. Inspired by its success, recent studies have explored adapting Mamba for multimodal fusion tasks. For instance, Pan-mamba~\cite{he2024pan} and Fusion-mamba~\cite{dong2024fusion} incorporate features from other modalities as inputs to unimodal Mamba to enable cross-modal interaction and fusion. Similarly, MambaDFuse~\cite{li2024mambadfuse} and MTMamba~\cite{lin2024mtmamba} utilize multimodal representations as inputs to unimodal Mamba for cross-modal interaction and fusion. In contrast, some approaches adopt a simpler strategy: VL-Mamba~\cite{qiao2024vl} and Cobra~\cite{zhao2024cobra}, for example, concatenate visual and textual representation sequences before feeding them into Mamba for sequence modeling and multimodal fusion.

While these Mamba-based approaches demonstrate significant computational advantages compared to Transformer-based multimodal fusion methods, they face inherent limitations due to Mamba's sequential scanning mechanism. This mechanism makes it challenging to effectively learn cross-modal correspondences, particularly with unscanned tokens. The resulting loss in cross-modal alignment information may constrain the effectiveness of learned multimodal fusion representations. Therefore, how to effectively leverage cross-modal relationships within the Mamba framework to learn more comprehensive multimodal fusion representations remains an open research challenge.
\section{Method}
\begin{figure*}[t]
\centering
\includegraphics[width=\linewidth]{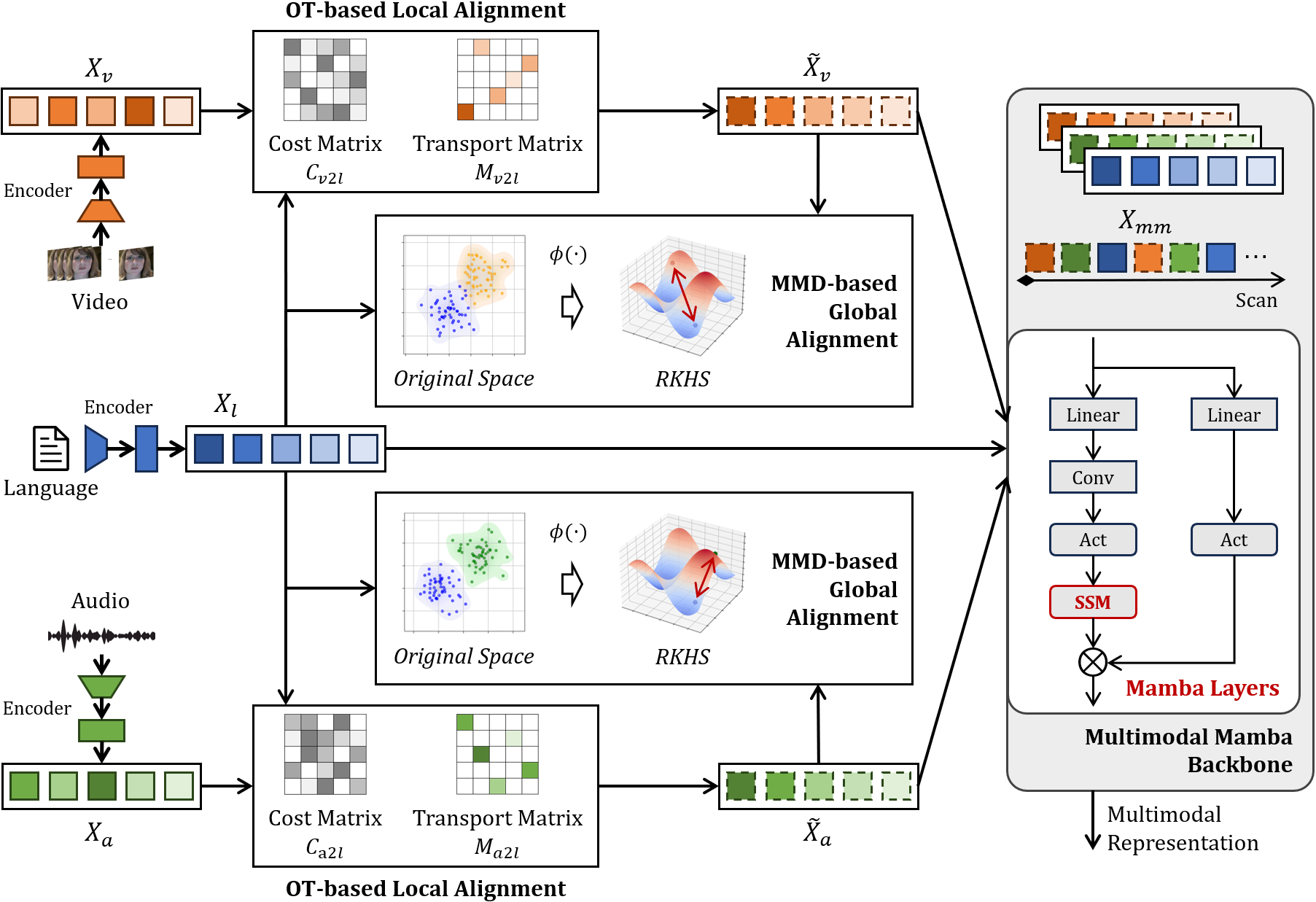}
\caption{AlignMamba enhances multimodal Mamba by incorporating token-level alignment and distribution-level alignment, enabling more effective multimodal fusion.}
\label{fig:framework}
\end{figure*}

\subsection{Overview}
Fig.~\ref{fig:framework} presents the framework of our proposed AlignMamba. Using audio-visual-language trimodal data as a case study, the framework first processes raw signals from each modality through modality-specific encoders to generate corresponding unimodal embedding sequences $X_a$, $X_v$, and $X_l$. The framework then employs two complementary alignment mechanisms: an OT-based local alignment module that captures token-level correspondences, and an MMD-based global alignment loss that ensures distribution-level consistency. These mechanisms yield aligned embedding sequences $\tilde{X}_a$ and $\tilde{X}_v$ (illustrated here by aligning audio and visual modalities to the language modality as the anchor). The aligned unimodal embeddings, which now incorporate cross-modal correspondence information, are subsequently processed by the Mamba backbone for multimodal fusion. The following sections provide a detailed description of each component.

\subsection{OT-based Local Cross-modal Alignment}
Optimal Transport provides a principled framework for comparing and aligning probability distributions by finding the optimal way to transform one distribution into another while minimizing the transportation cost~\cite{villani2021topics}. In our multimodal alignment context, OT offers a natural way to establish token-level correspondences between different modalities by treating feature sequences as discrete distributions.

Given the unimodal feature sequences $X_a \in \mathbb{R}^{T_a \times d}$, $X_v \in \mathbb{R}^{T_v \times d}$, and $X_l \in \mathbb{R}^{T_l \times d}$ from audio, video, and language modalities respectively, where $T_a$, $T_v$, and $T_l$ denote the sequence lengths of different modalities and $d$ is the feature dimension, we aim to learn the transport matrix $M$ that capture fine-grained correspondences between different modalities. Take video-to-language alignment as an example, the classical optimal transport problem can be formulated as follows:
\begin{equation}
\min_{T_{v2l}} \sum_{i=1}^{T_v}\sum_{j=1}^{T_l} M_{v2l}(i,j)C_{v2l}(i,j).
\end{equation}
The optimization is constrained by:
\begin{equation}
\begin{cases}
\sum_{j=1}^{T_l} M_{v2l}(i,j) = \frac{1}{T_v}, \quad \forall i \in [1,T_v] \\
\sum_{i=1}^{T_v} M_{v2l}(i,j) = \frac{1}{T_l}, \quad \forall j \in [1,T_l] \\
M_{v2l}(i,j) \geq 0, \quad \forall i,j
\end{cases}
\end{equation}
where $C_{v2l} \in \mathbb{R}^{T_v \times T_l}$ is the cost matrix. Given that the cosine distance emphasizes angular relationships between feature vectors while providing numerical stability through its bounded range, we use cosine distance as the cost matrix:
\begin{equation}
C_{v2l}(i,j) = 1 - \frac{X_v^i \cdot X_l^j}{||X_v^i||_2||X_l^j||_2}.
\end{equation}
However, solving this OT problem is extremely computationally expensive. Following~\cite{kusner2015word}, we adopt a relaxed version by removing the incoming sum constraint:
\begin{equation}
\begin{cases}
\sum_{j=1}^{T_l} M_{v2l}(i,j) = \frac{1}{T_v}, \quad \forall i \in [1,T_v] \\
M_{v2l}(i,j) \geq 0, \quad \forall i,j
\end{cases}
\end{equation}
This relaxed formulation allows each textual feature to be matched with multiple video features without constraining the total incoming flow, significantly reducing the computational complexity while maintaining the ability to capture meaningful cross-modal correspondences. The corresponding solution is defined as:
\begin{equation}
M_{v2l}(i,j) = \begin{cases}
\frac{1}{T_v}, & \quad j = \arg\min_j C_{v2l}(i,j), \\
0, & \quad j \neq \arg\min_j C_{v2l}(i,j).
\end{cases}
\end{equation}
Similarly, we compute the transport matrix $M_{a2l}$ for audio-to-language alignment. Finally, the aligned video and audio features can then be obtained through:
\begin{equation}
\begin{cases}
\begin{aligned}
\tilde{X}_v & = M_{v2l}^\top X_v \in \mathbb{R}^{T_l \times d}, \\
\tilde{X}_a & = M_{a2l}^\top X_a \in \mathbb{R}^{T_l \times d}.
\end{aligned}
\end{cases}
\end{equation}

This relaxed OT-based alignment process provides an efficient way to capture fine-grained cross-modal correspondences while maintaining computational tractability. The resulting transport matrices provide interpretable alignment information between different modalities. However, while this token-level alignment effectively captures local correspondences, ensuring global distribution-level consistency across modalities requires additional consideration, which we address through our MMD-based global alignment mechanism in the following section.

\subsection{MMD-based Global Cross-modal Alignment}
To ensure distribution-level consistency across modalities, we employ Maximum Mean Discrepancy as the global alignment metric. MMD measures the statistical discrepancy between different modalities in a high-dimensional Reproducing Kernel Hilbert Space (RKHS) by comparing all orders of their statistics. For two feature sequences $X$ and $Y$, the squared MMD distance is defined as:
\begin{equation}
\text{MMD}^2(X,Y) = \left\|\frac{1}{T}\sum_{i=1}^{T}\phi(x_i) - \frac{1}{T}\sum_{j=1}^{T}\phi(y_j)\right\|_{\mathcal{H}}^2,
\end{equation}
where $\phi(\cdot)$ is a feature mapping to a RKHS $\mathcal{H}$. Using the kernel trick, this can be computed as:
\begin{equation}
\begin{aligned}
\text{MMD}^2(X,Y) = & \frac{1}{T^2}\sum_{i=1}^{T}\sum_{i'=1}^{T}k(x_i,x_{i'}) + \frac{1}{T^2}\sum_{j=1}^{T}\sum_{j'=1}^{T}k(y_j,y_{j'}) \\
& - \frac{2}{T^2}\sum_{i=1}^{T}\sum_{j=1}^{T}k(x_i,y_j),
\end{aligned}
\end{equation}
where $k(\cdot,\cdot)$ is a positive definite kernel function. In our implementation, we adopt the Gaussian kernel:
\begin{equation}
k(x,y) = \exp(-\frac{\|x-y\|_2^2}{2\sigma^2}),
\end{equation}
where $\sigma$ is the kernel bandwidth parameter.

For the aligned audio features $\tilde{X}_a$, the aligned video features $\tilde{X}_v$, and the language features $X_l$, the global alignment loss is defined as the sum of MMD distances between each pair of modalities:
\begin{equation}
\mathcal{L}_{\text{align}} = \text{MMD}^2(\tilde{X}_v,X_l) + \text{MMD}^2(\tilde{X}_a,X_l).
\end{equation}

By minimizing this loss during training, we encourage the feature distributions of different modalities to be aligned in the RKHS. While OT establishes token-level correspondences, MMD ensures the consistency of overall feature distributions, providing complementary alignment signals at different granularities. This dual-alignment strategy facilitates more effective multimodal fusion in subsequent processing stages.

\subsection{Mamba-based Fusion and Optimization}
\noindent \textbf{Mamba-based Multimodal Fusion.} Following the local and global alignment processes, we employ Mamba to facilitate efficient multimodal fusion while maintaining its inherent linear computational complexity. Unlike traditional Transformer-based methods that process all tokens simultaneously through self-attention mechanisms, our approach implements a time-priority scanning strategy that preserves Mamba's sequential nature while enabling effective cross-modal interactions. Given the aligned audio features $\tilde{X}_a$, the aligned video features $\tilde{X}_v$, and the language features $X_l$, we construct a unified multimodal feature sequence $X_{mm}$ by interleaving features from different modalities at each timestep:
\begin{equation}
X_{mm} = [\tilde{X}_a^1, \tilde{X}_v^1, X_l^1, \tilde{X}_a^2, \tilde{X}_v^2, X_l^2, ..., \tilde{X}_a^T, \tilde{X}_v^T, X_l^T],
\end{equation}
where the superscript denotes the temporal index. This temporal-priority organization ensures that features from different modalities at the same timestep are processed sequentially, allowing the selective scan mechanism of Mamba to effectively capture both intra- and inter-modal dependencies. The fused representations are obtained by processing the constructed sequence through multiple Mamba layers.

\noindent \textbf{Training Objective.} The framework is optimized end-to-end using a composite loss function that combines the task-specific objective with the alignment constraints:
\begin{equation}
\mathcal{L} = \mathcal{L}_{task} + \lambda\mathcal{L}_{align},
\end{equation}
where $\mathcal{L}_{task}$ is determined by the downstream task (e.g., cross-entropy loss for classification or mean squared error for regression), $\mathcal{L}_{align}$ is the MMD-based alignment loss, and $\lambda$ is a hyperparameter that balances the two objectives. During training, minimizing $\mathcal{L}_{task}$ drives the model to learn task-relevant multimodal representations, while $\mathcal{L}_{align}$ ensures consistent feature distributions across modalities.
\section{Experiment}
We evaluate our proposed method on two distinct multimodal fusion scenarios: complete multimodal fusion and incomplete multimodal fusion. In the complete fusion setting, all modalities are available during both training and inference, which tests the model's ability to effectively integrate complementary information across modalities. The incomplete fusion scenario, where certain modalities may be missing during inference, presents a more challenging yet practical setting that evaluates the model's robustness and adaptability to partial observations. Through extensive experiments on these two scenarios, we demonstrate the effectiveness of our approach in both ideal conditions and more challenging practical situations.

\subsection{Datasets and Evaluation Metrics}
We conduct experiments on two multimodal representation fusion benchmarks: CMU-MOSI \cite{zadeh2016mosi} and CMU-MOSEI \cite{zadeh2018multimodal}. Both datasets consist of video segments collected from online platforms, containing visual (facial expressions), acoustic (voice), and textual (transcribed speech) modalities. Compared to CMU-MOSI, CMU-MOSEI exhibits greater diversity in terms of speakers, topics, and recording conditions. Each segment in both datasets is annotated with a sentiment score ranging from -3 (highly negative) to +3 (highly positive). These scores are binarized into positive and negative sentiments for classification. To evaluate the effectiveness of our method, we adopt the following metrics based on previous works~\cite{lian2023gcnet, wang2024incomplete}: binary accuracy and binary F$_1$ score.

\subsection{Comparison with SoTA methods}
\subsubsection{Results on Complete Multimodal Fusion Tasks}
\begin{table*}[htbp]
\centering
\begin{tabular}{c|c|c|c|c|c|c|c|c}
\toprule
Dataset & Missing & DCCA \cite{andrew2013deep} & DCCAE \cite{wang2015deep} & MCTN \cite{pham2019found} & MMIN \cite{zhao2021missing} & GCNet \cite{lian2023gcnet} & IMDer \cite{wang2024incomplete} & \textbf{AlignMamba} \\ 
\midrule
\multirow{9}{*}{MOSI} 
& 10\% & 72.1 / 72.2 & 74.5 / 74.7 & 78.4 / 78.5 & 81.8 / 81.8 & 82.3 / 82.3 & 84.9 / 84.8 & \textbf{85.7 / 85.6} \\ 
& 20\% & 69.3 / 69.1 & 71.8 / 71.9 & 75.6 / 75.7 & 79.0 / 79.1 & 79.4 / 79.5 & 83.5 / 83.4 & \textbf{84.3 / 84.1} \\ 
& 30\% & 65.4 / 65.2 & 67.0 / 66.7 & 71.3 / 71.2 & 76.1 / 76.2 & 77.2 / 77.2 & 81.2 / 81.0 & \textbf{82.2 / 82.2} \\ 
& 40\% & 62.8 / 62.0 & 63.6 / 62.8 & 68.0 / 67.6 & 71.7 / 71.6 & 74.3 / 74.4 & 78.6 / 78.5 & \textbf{80.0 / 79.6} \\ 
& 50\% & 60.9 / 59.9 & 62.0 / 61.3 & 65.4 / 64.8 & 67.2 / 66.5 & 70.0 / 69.8 & 76.2 / 75.9 & \textbf{77.6 / 77.3} \\ 
& 60\% & 58.6 / 57.3 & 59.6 / 58.5 & 63.8 / 62.5 & 64.9 / 64.0 & 67.7 / 66.7 & 74.7 / 74.0 & \textbf{75.8 / 75.1} \\ 
& 70\% & 57.4 / 56.0 & 58.1 / 57.4 & 61.2 / 59.0 & 62.8 / 61.0 & 65.7 / 65.4 & 71.9 / 71.2 & \textbf{73.8 / 73.2} \\
& Avg. & 63.8 / 63.1 & 65.2 / 64.8 & 69.1 / 68.5 & 71.9 / 71.5 & 73.8 / 73.6 & 78.7 / 78.4 & \textbf{79.9 / 79.6} \\ 
& $\Delta$ & 14.7 / 16.2 & 16.4 / 17.3 & 17.2 / 19.5 & 19.0 / 20.8 & 16.6 / 16.9 & 13.0 / 13.6 & \textbf{11.9 / 12.4} \\
\midrule
\multirow{9}{*}{MOSEI} 
& 10\% & 77.4 / 77.3 & 78.4 / 78.3 & 81.8 / 81.6 & 81.9 / 81.3 & 82.3 / 82.1 & 84.8 / 84.6 & \textbf{85.4 / 85.4} \\ 
& 20\% & 73.8 / 74.0 & 75.5 / 75.4 & 79.0 / 78.7 & 79.8 / 78.8 & 80.3 / 79.9 & 82.7 / 82.4 & \textbf{83.6 / 83.3} \\ 
& 30\% & 71.1 / 71.2 & 72.3 / 72.2 & 76.9 / 76.2 & 77.2 / 75.5 & 77.5 / 76.8 & 81.3 / 80.7 & \textbf{82.5 / 81.0} \\ 
& 40\% & 69.5 / 69.4 & 70.3 / 70.0 & 74.3 / 74.1 & 75.2 / 72.6 & 76.0 / 74.9 & 79.3 / 78.1 & \textbf{81.7 / 80.5} \\ 
& 50\% & 67.5 / 65.4 & 69.2 / 66.4 & 73.6 / 72.6 & 73.9 / 70.7 & 74.9 / 73.2 & 79.0 / 77.4 & \textbf{80.1 / 78.7} \\ 
& 60\% & 66.2 / 63.1 & 67.6 / 63.2 & 73.2 / 71.1 & 73.2 / 70.3 & 74.1 / 72.1 & 78.0 / 75.5 & \textbf{79.4 / 78.2} \\ 
& 70\% & 65.6 / 61.0 & 66.6 / 62.6 & 72.7 / 70.5 & 73.1 / 69.5 & 73.2 / 70.4 & 77.3 / 74.6 & \textbf{78.8 / 76.9} \\ 
& Avg. & 70.2 / 68.8 & 71.4 / 69.7 & 75.9 / 75.0 & 76.3 / 74.1 & 76.9 / 75.6 & 80.3 / 79.0 & \textbf{81.6 / 80.6} \\
& $\Delta$ & 11.8 / 16.3 & 11.8 / 15.7 & 9.1 / 11.1 & 8.8 / 11.8 & 9.1 / 11.7 & 7.5 / 10.0 & \textbf{6.6 / 8.5} \\
\bottomrule
\end{tabular}
\caption{Performance comparison on CMU-MOSI and CMU-MOSEI datasets. Results are reported as Accuracy / F$_1$ (\%). $\Delta$: performance drop from 10\% to 70\% missing rate (lower is better).}
\label{tab:incomplete}
\end{table*}

\begin{table}[htbp]
\centering
\begin{tabular}{c|c|c}
\toprule
Method & CMU-MOSI & CMU-MOSEI \\
\midrule
ICCN \cite{sun2020learning} & 83.0 / 83.0 & 84.2 / 84.2  \\
MISA \cite{hazarika2020misa} & 83.4 / 83.6 & 85.5 / 85.3  \\
MulT \cite{tsai2019multimodal} & 84.1 / 83.9 & 82.5 / 82.3 \\
MAG-BERT \cite{rahman2020integrating} & 84.3 / 84.6 & 84.8 / 84.7  \\
CM-BERT \cite{yang2020cm} & 84.5 / 84.5 & 83.6 / 83.6 \\
ULGM \cite{hwang2023self} & 84.5 / 84.5 & 85.0 / 85.1 \\
FDMER \cite{yang2022disentangled} & 84.6 / 84.7 & 86.1 / 85.8 \\
Self-MM \cite{yu2021learning} & 84.8 / 84.9 & 85.0 / 84.9 \\
MMIM \cite{han2021improving} & 85.1 / 85.0 & 85.1 / 85.0 \\
HyCon \cite{mai2022hybrid} & 85.2 / 85.1 & 85.4 / 85.6 \\
Confede \cite{yang2023confede} & 85.5 / 85.5 & 85.8 / 85.8 \\
AOBERT \cite{kim2023aobert} & 85.6 / 86.4 & 86.2 / 85.9 \\
DMD \cite{li2023decoupled} & 85.8 / 85.8 & 86.0 / 86.1 \\
MTMD \cite{lin2023multi} & 86.0 / 86.0 & 86.1 / 85.9 \\
\midrule
\textbf{AlignMamba} & \textbf{86.9 / 86.9} & \textbf{86.6 / 86.5} \\
\bottomrule
\end{tabular}
\caption{Performance comparison on CMU-MOSI and CMU-MOSEI datasets. Results are reported as Accuracy / F$_1$ (\%).}
\label{tab:complete}
\end{table}

Table \ref{tab:complete} presents a comprehensive comparison between our approach and various state-of-the-art methods on the complete multimodal representation fusion task, which can be categorized into three main groups: (1) LSTM methods, including ICCN \cite{sun2020learning}, MISA \cite{hazarika2020misa}, and MMIM \cite{han2021improving}; (2) Cross-modal Transformer methods: MulT \cite{tsai2019multimodal}, Self-MM \cite{yu2021learning}, and DMD \cite{li2023decoupled}; (3) Contrastive learning methods: HyCon \cite{mai2022hybrid}, Confede \cite{yang2023confede}, and MTMD \cite{lin2023multi}.

On one hand, AlignMamba additionally incorporates token-level alignment to enhance multimodal fusion compared to contrastive learning approaches. On the other hand, AlignMamba's advantage over cross-modal Transformer methods lies in its consideration of distributional alignment relationships. Consequently, AlignMamba attains the best performance on all metrics in both datasets. For example, on the CMU-MOSI dataset, AlignMamba achieves a binary classification accuracy of 86.9\%, representing a 0.9\% improvement over previous methods. These results can be ascribed to AlignMamba's capacity to conduct extensive cross-modal alignment by leveraging its local alignment module and global alignment loss, thereby adeptly exploiting cross-modal correlations across different granularities and enabling the learning of more effective multimodal fusion representations.

\subsubsection{Results on Incomplete Multimodal Fusion Tasks}
Table \ref{tab:incomplete} presents experimental results on incomplete multimodal representation fusion tasks. We compare AlignMamba with various state-of-the-art methods, which can be categorized into two main groups: (1) modality recovery approaches, including MCTN \cite{pham2019found}, MMIN \cite{zhao2021missing}, GCNet \cite{lian2023gcnet}, and IMDer \cite{wang2024incomplete}, which attempt to reconstruct missing modalities from available ones; and (2) non-recovery approaches, such as DCCA \cite{andrew2013deep} and DCCAE \cite{wang2015deep}, which directly learn from available modalities.

The results demonstrate that AlignMamba consistently outperforms existing methods across different missing rates, achieving an average accuracy of 79.9\% on the CMU-MOSI dataset, a 1.2\% improvement over previous methods. More importantly, AlignMamba demonstrates stronger robustness to increasing modality missing rates. For instance, on the CMU-MOSI dataset, while MMIN and IMDer experience significant performance degradation with accuracy drops of 19.0\% and 13.0\% respectively, AlignMamba shows better resilience with only an 11.9\% decrease in binary classification accuracy.

In conclusion, these improvements on both complete and incomplete multimodal fusion tasks can be attributed to the proposed dual alignment strategy: the local token-level alignment and global distribution-level alignment mechanisms work together to capture comprehensive cross-modal correspondences. This dual alignment strategy, combined with Mamba's efficient sequence modeling capabilities, not only enables learning more comprehensive and accurate multimodal fusion representations in complete multimodal scenarios, but also improves the robustness of learned representations in incomplete multimodal settings.

\subsection{Efficiency Analysis}
We conduct comprehensive efficiency analysis for AlignMamba and compare them against both single-stream and multi-stream Transformer methods. Our evaluation metrics consist of GPU memory usage, inference time, and computational complexity. For a fair comparison, we specifically focus on the cross-modal interaction and fusion components, excluding the computational costs of unimodal encoders. All experiments are performed under identical conditions.

\subsubsection{GPU Memory Usage}
\begin{figure}[htbp]
\centering
\includegraphics[width=\linewidth]{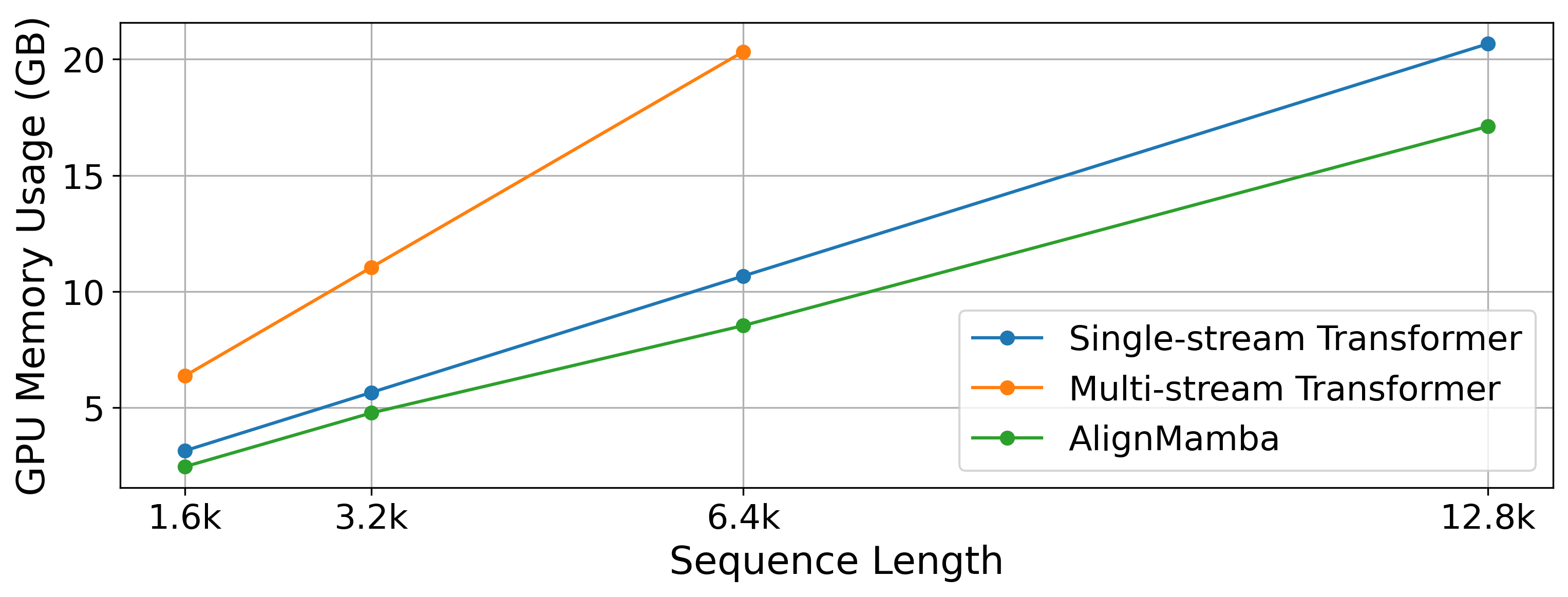}
\caption{GPU memory usage comparison with varying lengths.}
\label{fig:gpu}
\end{figure}

First, we report the GPU memory usage of each method with respect to varying input sequence lengths in Fig.~\ref{fig:gpu}. We exclude multi-stream Transformers on the 12.8k-token setting as they encounter an out-of-memory error. AlignMamba consistently achieves the best trade-off between sequence length and memory usage across all settings, surpassing other Transformer-based approaches by a non-trivial margin. For instance, when processing 6.4k tokens, AlignMamba requires only 8.53 GB of memory, achieving 20.3\% and 58.0\% memory reduction compared to single-stream (10.7 GB) and multi-stream (20.3 GB) Transformers, respectively. This significant advantage in memory consumption is particularly valuable for processing longer sequences and deploying models on resource-constrained devices.

\subsubsection{Inference Time}
\begin{figure}[htbp]
\centering
\includegraphics[width=\linewidth]{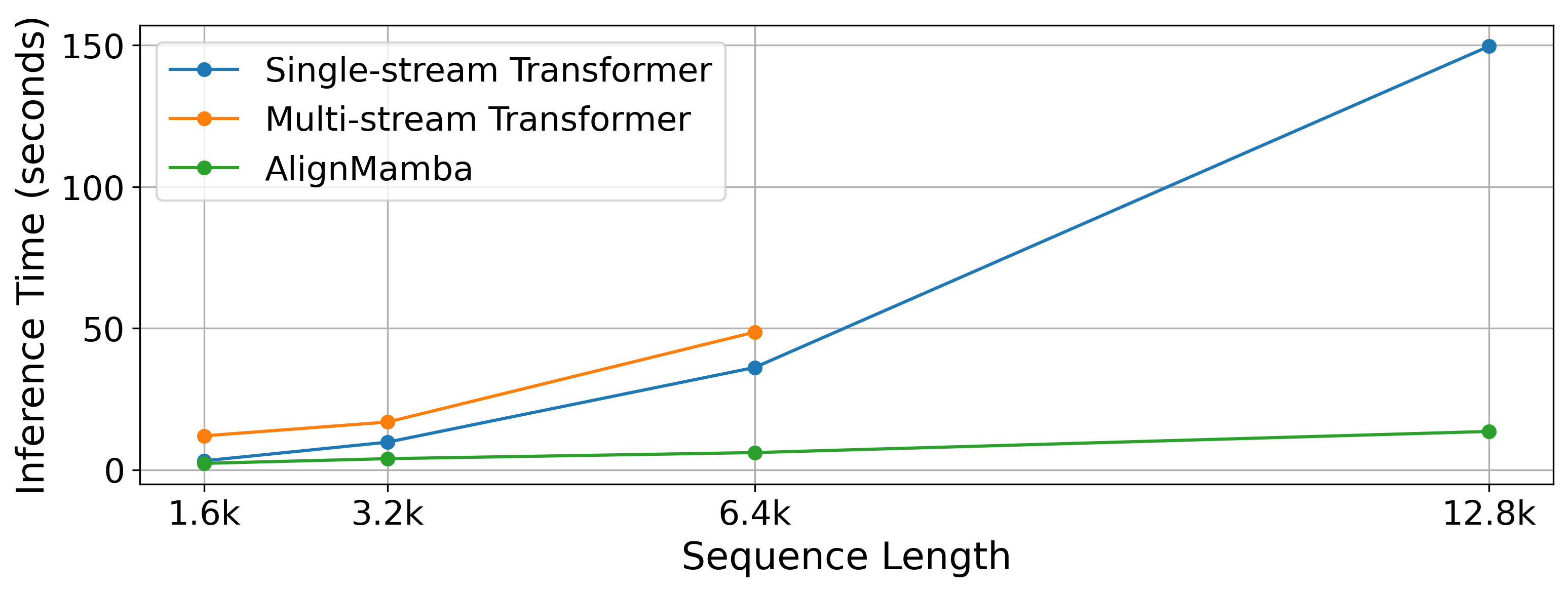}
\caption{Inference time comparison with varying lengths.}
\label{fig:inference}
\end{figure}

Next, we report the inference time of each method with respect to varying input sequence lengths in Fig.~\ref{fig:inference}. To ensure fairness, we aggregate the running time of 50 inference passes for each model. AlignMamba again demonstrates consistent and substantial speed advantages over Transformer-based approaches across all settings. For instance, when processing 6.4k tokens, AlignMamba takes only 6.05 seconds, achieving 83.3\% and 87.6\% reduction in inference time compared to single-stream (36.13s) and multi-stream (48.61s) Transformers, respectively.

\subsubsection{Computational Complexity}
Finally, we analyze the FLOPs required by each method to quantify their computational efficiency. Without loss of generality, we fix the input sequence length to 1024 for each model. AlignMamba demonstrates superior efficiency with only 46.7G FLOPs, compared to 101.6G FLOPs for single-stream Transformer and 203.2G FLOPs for multi-stream Transformer. This represents a reduction of more than 54\% compared to single-stream and 77\% compared to multi-stream approaches, highlighting AlignMamba's computational advantages in cross-modal alignment and multimodal fusion tasks. This also justifies the lower memory consumption and swift inference speed as presented in the previous sections.

\begin{figure*}[htbp]
\centering
\includegraphics[width=\linewidth]{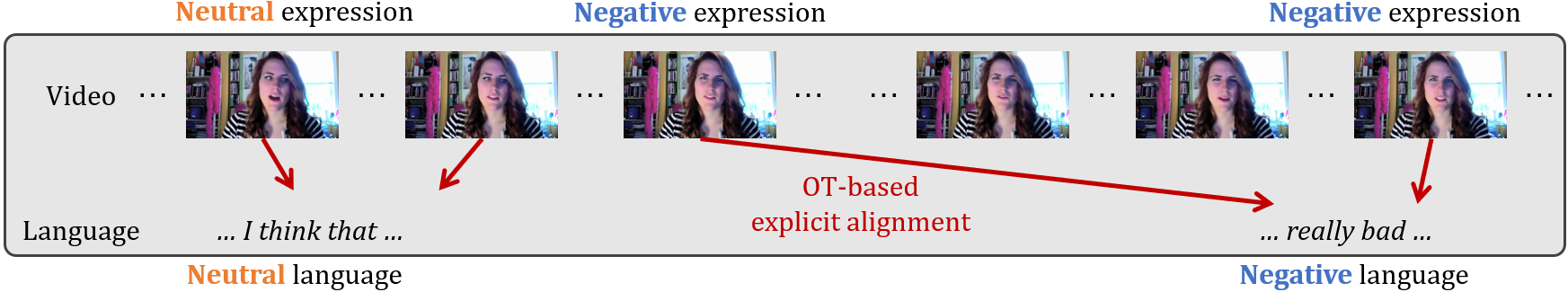}
\caption{The learned optimal transport plan. We only show the transport plan between video and language modalities for brevity.}
\label{fig:ot}
\end{figure*}

\subsection{Ablation study}
We conduct comprehensive ablation studies from three aspects to evaluate our proposed method, as shown in Table \ref{tab:ablation}.

\noindent \textbf{Component analysis.} First, we evaluate the effectiveness of the OT-based local alignment module and MMD-based global alignment loss. Removing either component led to performance degradation across both datasets. For instance, on the CMU-MOSI dataset, accuracy dropped by 2.3\% and 1.1\% respectively. Notably, the OT-based alignment module demonstrated superior performance compared to the MMD-based alignment loss, likely because OT-based alignment provides explicit alignment plans while MMD-based alignment only imposes implicit alignment constraints.

\noindent \textbf{Mamba-based fusion.} Furthermore, we ablate AlignMamba with regular single-stream \cite{zhao2024cobra} and multi-stream Mamba-based fusion methods \cite{dong2024fusion} to show the effectiveness of our method in terms of multimodal fusion. Results demonstrate reduced performances in these two Mamba-based methods, suggesting their lack of explicit consideration of inter-modal correspondences, which makes it difficult to learn comprehensive cross-modal relationships. This shows that naive Mamba architecture alone does not suffice in effective multimodal fusion and highlights both the limitations of Mamba's original scanning mechanism and the necessity of our proposed cross-modal alignment.

\noindent \textbf{Modality ablations.} Lastly, we conduct modality ablation experiments by removing one modality at a time. When the text modality is removed, we only align the audio modality with the video modality. This results in significant performance degradation, likely due to the strong correlation between language and emotions. In contrast, removing the audio modality results in a smaller performance drop, possibly due to the sizable presence of irrelevant information in the audio modality such as background noise, reducing its impact on the overall performance.

\begin{table}[htbp]
\centering
\begin{tabular}{ccc}
\toprule
 & CMU-MOSI & CMU-MOSEI \\
\midrule
\textbf{AlignMamba} & \textbf{86.9 / 86.9} & \textbf{86.6 / 86.5} \\
\midrule
\multicolumn{3}{c}{Alignment} \\
\midrule
w/o Local & 84.6 / 84.4 & 84.1 / 84.0 \\
w/o Global & 85.8 / 85.7 & 85.7 / 85.5 \\
\midrule
\multicolumn{3}{c}{Fusion} \\
\midrule
Single-stream & 82.3 / 82.1 & 81.8 / 81.4 \\
Multi-stream & 83.7 / 83.5 & 83.5 / 83.2 \\
\midrule
\multicolumn{3}{c}{Modality} \\
\midrule
w/o Audio & 84.4 / 84.6 & 83.9 / 83.5 \\
w/o Video & 83.7 / 83.8 & 83.3 / 82.8 \\
w/o Language & 65.3 / 63.4 & 64.6 / 62.2 \\
\bottomrule
\end{tabular}
\caption{Ablation studies on CMU-MOSI and CMU-MOSEI datasets. Results are reported as Accuracy / F$_1$ (\%).}
\label{tab:ablation}
\end{table}

\subsection{Further Analysis}
\subsubsection{Cross-modal Alignment}
To quantitatively assess our dual alignment strategy, we measure the $\mathcal{A}$-distance between modality pairs in Table~\ref{tab:distance}. The $\mathcal{A}$-distance $\in [0, 2]$ is a common metric for domain discrepancy, with higher values indicating greater modality differences. $\mathcal{A}_{al}$ and $\mathcal{A}_{vl}$ represents the audio-language and video-language distances respectively. The results reveal significant and consistent reductions in inter-modal distances through our dual alignment strategy in both CMU-MOSI and CMU-MOSEI. These improvements demonstrate the effectiveness of our strategy in bridging the modality gap by learning meaningful cross-modal correlations, leading to more robust multimodal fusion representations.

\begin{table}[h]
\centering
\begin{tabular}{c|cc|cc}
\toprule
 & \multicolumn{2}{c}{CMU-MOSI} & \multicolumn{2}{c}{CMU-MOSEI} \\
 & $\mathcal{A}_{al}$ & $\mathcal{A}_{vl}$ & $\mathcal{A}_{al}$ & $\mathcal{A}_{vl}$ \\
\midrule
w/ Dual-align & 1.59 & 1.49 & 1.61 & 1.53 \\
w/o Dual-align & 1.68 & 1.57 & 1.72 & 1.65 \\
\bottomrule
\end{tabular}
\caption{$\mathcal{A}$-distance between different modalities.}
\label{tab:distance}
\end{table}

\subsubsection{Optimal Transport Plan}
Here, we qualitatively present the learned optimal transport plan. Figure \ref{fig:ot} illustrates an example from the CMU-MOSI dataset. Notice that modalities exhibit temporal misalignment: the sentiment correspondence between different modalities may appear at different timesteps, which poses a challenge for multimodal representation fusion. For instance, the visual modality exhibits negative expressions at the beginning, while the textual modality introduces negative words toward the end. The original Mamba model struggles to explicitly learn these correspondences due to its sequential scanning mechanisms. In contrast, our proposed method leverages optimal transport to explicitly transform and align features in different temporal stages across modalities, reducing the modality gap and improving the effectiveness of multimodal fusion.

\section{Conclusion}
In this paper, we proposed AlignMamba, an efficient and effective method for multimodal representation fusion. By integrating OT-based local alignment and MMD-based global alignment, our method captures comprehensive cross-modal relationships while maintaining lower computational complexity. Extensive experiments on both complete and incomplete multimodal fusion tasks demonstrate that AlignMamba achieves state-of-the-art performance with significantly reduced computational costs.
{
    \small
    \bibliographystyle{ieeenat_fullname}
    \bibliography{main}
}

\end{document}